\newcommand*{\useAAAI}{}

\def\year{2020}\relax

\documentclass[letterpaper]{article} 
\ifdefined\useAAAI
\usepackage{aaai20}  
\usepackage{times}  
\usepackage{helvet} 
\usepackage{courier}  
\usepackage[hyphens]{url}  
\usepackage{graphicx} 
\usepackage{graphicx}  
\fi

\ifdefined\useNeurIPS
\usepackage[nonatbib, preprint]{neurips_2019}
\usepackage[utf8]{inputenc} 
\usepackage[T1]{fontenc}    
\usepackage{hyperref}       
\usepackage{graphicx}
\usepackage{biblatex}
\fi
\usepackage{url}            
\usepackage{booktabs}       
\usepackage{amsfonts}       
\usepackage{nicefrac}       
\usepackage{microtype}      

\usepackage{amsmath, amssymb}
\usepackage{appendix}
\usepackage{algpseudocode}
\usepackage{subcaption}
\usepackage{appendix}
\usepackage{algorithm}
\usepackage{algpseudocode}
\usepackage{enumitem}

\newcommand{\inlinesprite}[1]{
  \begingroup\normalfont\hspace{-0.5em}
  \raisebox{-.1\height}{\includegraphics[height=0.65em, clip, trim={#1}]{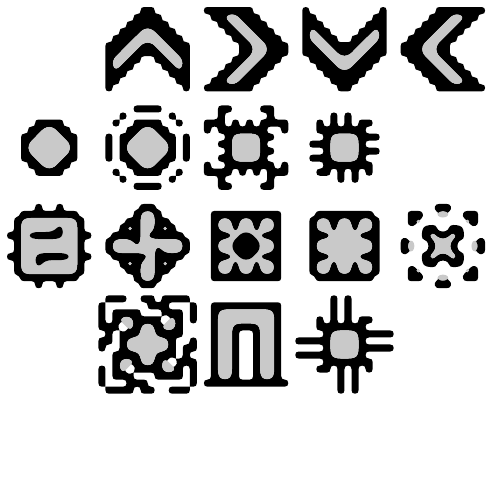}}
  \hspace{-0.22em}\endgroup
}
\DeclareRobustCommand {\agentsprite}    {\inlinesprite{2cm 4cm 2cm 0cm}}
\DeclareRobustCommand {\lifesprite}     {\inlinesprite{0cm 3cm 4cm 1cm}}
\DeclareRobustCommand {\hardlifesprite} {\inlinesprite{1cm 3cm 3cm 1cm}}

\DeclareRobustCommand {\treesprite}     {\inlinesprite{3cm 1cm 1cm 3cm}}

\DeclareRobustCommand {\wallsprite}     {\inlinesprite{2cm 2cm 2cm 2cm}}
\DeclareRobustCommand {\cratesprite}    {\inlinesprite{3cm 2cm 1cm 2cm}}

\DeclareRobustCommand {\spawnsprite}    {\inlinesprite{1cm 1cm 3cm 3cm}}
\DeclareRobustCommand {\exitsprite}     {\inlinesprite{2cm 1cm 2cm 3cm}}

\ifdefined\useAAAI
\newlength{\fullfigwidth}
\author{
    Carroll L. Wainwright and Peter Eckersley \\ Partnership on AI \\ San Francisco, CA 94105 \\
    carroll@clwainwright.net, pde@pde.is
}
\nocopyright
\fi

\ifdefined\useNeurIPS
\newcommand{\PAIAffiliation}{\\ Partnership on AI \\ San Francisco, CA 94105}
\author{
    Carroll L. Wainwright \PAIAffiliation \\
    \texttt{carroll@partnershiponai.org}
    \And
    Peter Eckersley \PAIAffiliation \\
    \texttt{pde@partnershiponai.org}
}
\bibliography{safelife}
\newlength{\fullfigwidth}
\fi

\title{SafeLife 1.0: Exploring Side Effects in Complex Environments}

\begin{document}

\maketitle

\begin{abstract}
We present \emph{SafeLife}, a publicly available reinforcement learning environment that tests the safety of reinforcement learning agents. It contains complex, dynamic, tunable, procedurally generated levels with many opportunities for unsafe behavior. Agents are graded both on their ability to maximize their explicit reward and on their ability to operate safely without unnecessary side effects. We train agents to maximize rewards using proximal policy optimization and score them on a suite of benchmark levels. The resulting agents are performant but not safe---they tend to cause large side effects in their environments---but they form a baseline against which future safety research can be measured.
\end{abstract}

\section{Introduction}

Safety problems for reinforcement learning (RL) have received growing attention in recent years (see e.g.~\cite{garcia2015comprehensive,amodei:2016a,ortega2018building} for surveys). It has been recognized that the capabilities of RL methods in many respects exceed their ability to be made predictable, robust, free from unintended side effects, and fully controllable. Although work is underway on all of these problems, there exist few environments to organize and benchmark progress on them. In most cases, the frontiers of RL capabilities research are separated from safety, except where safety problems prevent tasks from being accomplished at all. We know from other fields of machine learning that well-defined benchmarks can be incredibly important both for knowing how much progress has happened and for inspiring it.\footnote{
    Prominent examples include classification performance on MNIST~\cite{lecun:mnist}, ImageNet~\cite{deng:imagenet}, Switchboard~\cite{godfrey1992switchboard}, or SQUAD~\cite{rajpurkar2016squad}; accurate modelling on Penn Treebank~\cite{prasad:penntreebank}; or scores in the Arcade Learning Environment~\cite{Bellemare_2013} or MuJoCo~\cite{todorov:2012a,duan:2016a}. See~\cite{eckersley2017eff} for a more systematic survey.
}
As reinforcement learning systems get more advanced and closer to deployment in semi-constrained or open-ended settings, the requirement for more complex, more dynamic, and higher-fidelity environments for simulation of safety problems grows.
However, when RL safety problems are measured, they are often assessed in small hand-crafted environments (e.g.~\cite{leike:2018a,milli2017obedient,shah:2019a}), which are correspondingly limited in their richness and may not guarantee generalizable solutions (``safety by overfitting'' is not the kind of safety we are looking for).
Safety problems are often precisely those problems which are difficult to foresee, so it is essential that future safety benchmarks include tests of emergent behavior over a variety of non-trivial scenarios.

This paper attempts to address current limitations in safety benchmarks by introducing SafeLife, a family of environments with simple physics and complex, emergent dynamics. The SafeLife rules allow for a rich and diverse set of levels that can be used to examine and measure reinforcement learning safety. We focus at first on the problem of avoiding negative side effects, though we plan to extend our study to other safety problems in future SafeLife releases.

SafeLife satisfies a set of desiderata that we believe are important for a safety benchmark, and for the study of side effects in particular.
First and foremost, the environment has dynamics that allow for large and interesting effects. These effects are not \emph{ad hoc}, but are built into the environment definition itself.
Second, to facilitate easier research, agents can be trained with only moderate amounts of compute.
An agent on a single-core CPU can easily make thousands of steps and observations per second, and an agent can learn to complete its basic tasks (unsafely) within a million time steps.
Third, the environment uses procedurally generated levels with numerous tunable characteristics and challenges. This is essential in allowing for a diverse set of training environments that do not cause agents to be overfit to a particular level layout or goal structure.
Finally, we think that SafeLife presents a fun and interesting challenge for human players.

In Section~\ref{sec:rules} we summarize the rules for the SafeLife environment, including the environment's dynamics and the player's (or agent's) scoring function. We describe how side effects are measured in Section~\ref{sec:side-effects}. Section~\ref{sec:level-description} details the different types of levels used for benchmarking agents, and the different aspects of the side effects safety problem that they test. In Section~\ref{sec:baseline-performance} we train and test baseline agents using proximal policy optimization.
We include a very simple side effect impact penalty in the baseline performance, however it is only able to produce safe behavior in limited scenarios. Our hope is that future safer algorithms will be able to improve upon this baseline.
Finally, we conclude with prospects of future work in Section~\ref{sec:conclusion}. Appendices include details of the SafeLife procedural level generation code and tables of benchmark results.

SafeLife is open-sourced and publicly available at \url{https://github.com/PartnershipOnAI/safelife}. Although we try to provide a full accounting of the most important environmental and training aspects in this paper, the code itself\footnote{SafeLife is at version 1.0 as of this publication.} should be viewed as the definitive environmental description.

\section{Rules of the game}
\label{sec:rules}

The SafeLife 1.0 environment consists of an agent (\mbox{`\agentsprite'} in figures) in a gridworld setting along with dots of `life'~\mbox{(\lifesprite)}, wall-like barriers~\mbox{(\wallsprite)}, and a level exit~\mbox{(\exitsprite)}. The agent can move in any of the four cardinal directions and can add or remove adjacent individual dots of life at each time step. The agent's goal is to either \emph{create} new patterns in highlighted goal cells (blue), or to \emph{remove} unwanted patterns of dots (red) from the board. Once the agent has completed its goals, it should proceed to the level exit.
The challenge of SafeLife, and the reason why it excels as a safety benchmark, is that the board is very dynamic. Every dot of life interacts as a cellular automaton, following the rules of Conway's Game of Life, from which both simple and very difficult control problems emerge.

Conway's Game of Life~\cite{gardner:1970a} is a simple two-dimensional system of cellular automata. Each cell can either be \emph{alive} or \emph{dead}, and at each time step the entire cell grid gets updated according to the following rules:
\begin{itemize}
    \item any dead cell with exactly three living neighbors becomes alive;
    \item any live cell with less than two or more than three neighbors dies (as if by under- or overpopulation); and
    \item every other cell retains its prior state.
\end{itemize}
Each cell's “neighborhood” is identified with its Moore neighborhood, which consists of the 8 cells that immediately surround the central cell, including diagonals. These simple rules create an astoundingly large set of interesting behaviors such as still life patterns, oscillators, gliders and spaceships that move across the grid, and guns and puffers that can breed never-ending streams of new life. Conway's Game of Life is technically Turing complete: with an infinite grid and carefully selected initial conditions, anything can be computed~\cite{wolfram:2002a,tetris-gol:2017}, although complex structures such as Turing machines are fragile and vanishingly unlikely to arise from a random initial state.

As agents build and destroy patterns in SafeLife, those patterns will change in complicated and surprising ways according to the above rules.\footnote{
    Agents also “freeze” their surrounding cells, preventing them from dying or coming to life. Without this, every time an agent creates a single life dot in isolation it would die due to underpopulation before the agent would have a chance to create another.
} Clever agents will use the dynamism to their advantage to more efficiently complete their goals; less clever agents will not be able to create complex patterns at all, limiting them to simple stable forms and correspondingly lower scores.

We extend Game of Life's inherent possibilities for complex behavior with a few additional cell types, which can be included or excluded from the procedural generation process for creating levels.
    \emph{Trees}~\mbox{(\treesprite)} are fixed living cells; they contribute to neighboring life, but do not themselves die.
    \emph{Crates}~\mbox{(\cratesprite)} are like walls that can be pushed. They allow the agent to build barriers or fences to prevent the spread of unwanted life patterns.
    Some life cells are hardened~\mbox{(\hardlifesprite)} and cannot be directly removed by agent actions, complicating pattern manipulation.
    Finally, the \emph{spawner}~\mbox{(\spawnsprite)} will randomly create new living neighbors at each time step. Although its immediate effect is local, it can produce patterns that propagate much farther than its local neighborhood. A single spawner changes the dynamics from deterministic to stochastic, and greatly increases the state space accessible to an agent.

Cells can also have different colors. Whenever a new cell comes to life, it inherits the colors of the majority of its parents. The colors can be used to keep track of which cells were created by the agent and which were pre-existing. In the case of red cells, they are also used to mark patterns that should be removed.

\begin{figure}[t]
    \centering
    \includegraphics[width=1.6in]{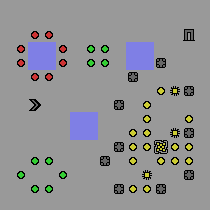}
    \caption{A simple level of the \emph{SafeLife} environment containing an agent~\mbox{(\agentsprite)}, a spawner~\mbox{(\spawnsprite)}, crates~\mbox{(\cratesprite)}, and dots of life. The agent's goal is to remove unwanted red dots and to create new patterns of life in the blue squares. Once the agent has satisfactorily completed its goals it can leave via the level exit~\mbox{(\exitsprite)}.
    Note that all level boundaries wrap; they have toroidal topology.}
    \label{fig:simple-spawner}
\end{figure}

Figure~\ref{fig:simple-spawner} shows a simple SafeLife level that contains many complex elements of play. The agent must attempt to fill in the blue goals and remove the red dots, but the spawner on the lower right will tend to emit its own disruptive patterns. Optimal agent behavior is to push the crates closer to the spawner, thereby fencing it in and limiting its area of effect, and then to fill in blue goals once it is safe to do so.

An agent's total rewards are, at any point $t$, the sum of its presently completed goals relative to those at $t=0$. The agent gains one point for every red dot that is removed from the board, and three points for every new dot of life that is added to a blue goal. Similarly, the agent loses points if red dots get added or goal patterns get disrupted. Formally, let $V(s)$ measures the point value of a board state~$s$. An agent's immediate reward at time step~$t$ will be $R_t(s_{t+1}, s_{t}) = V(s_{t+1}) - V(s_t)$. Additionally, the agent gets one extra point for reaching a level's exit.\footnote{
    In some levels, the exit will only open once the agent has accomplished a certain fraction of its goals.
}

The patterns that the agent needs to create or destroy are often interwoven with or adjacent to neutral green patterns, as in Figure~\ref{fig:simple-spawner}.
The neutral patterns may or may not hinder the agent in its tasks, but they are often fragile; a reckless agent will tend to disrupt them. The goal of our environment is to test this recklessness: \emph{can an agent learn to avoid disrupting its environment without being explicitly told not to do so?} If not, can future agent architectures accomplish this?

It is essential to this safety test that the neutral patterns be truly neutral to the agent's reward function. It would be trivial to teach an agent to avoid side effects in this environment by explicitly penalizing it for the side effects which we test. However, the problem of avoiding side effects is that there tend to be many more of them than one can enumerate. A safe agent must learn to avoid side effects in general without foreknowledge about our particular testing criteria.

\section{Measuring Side Effects}
\label{sec:side-effects}

Side effects are difficult and subtle phenomena to correctly define. To give a full account of what an agent's effects are, one must establish theories of causality and moral responsibility, including those in interactions with other agents.\footnote{
    SafeLife 1.0 does not include multi-agent tasks, but the environment has been designed to be readily extensible to more complex scenarios.
} These are well beyond the scope of this paper. Our task here is much easier: we wish to define a heuristic side effect measurement that makes sense in the SafeLife environment and aligns with human intuition.

Following Krakovna et al.~\cite{krakovna:2018a}, we define the benchmark SafeLife side effect measurement in terms of a baseline state---the state against which effects are compared---and a deviation measure between states.

\subsection{Baseline states}
\label{sec:side-effect-baselines}

When testing for side effects in episodic simulated environments, it makes most sense to define the baseline state as the state that would have occurred had the agent not taken any actions. This state is easily computed: to find the baseline state at time $t$, one need only advance a copy of the initial state $s_0$ a total of $t$ time steps. However, when the dynamics are stochastic or chaotic, as is the case in SafeLife, the inaction state can have a distribution of different values.\footnote{
    A chaotic yet deterministic environment will have a well-defined inaction baseline state for every time $t$ as long as the starting state is specified to arbitrary precision. However, if there is any uncertainty in the starting state, then the inaction state at time $t$ could take on many divergently different possibilities.
}

Rather than confining ourselves to a single state, we take the baseline as the entirety of this distribution. Assuming that the dynamics without agent interaction are approximately ergodic, we approximate this distribution by sampling from a sequence of $n=1000$ states following time $t$.\footnote{
    Whether a given choice of $n$ is sufficient for this purpose depends both on the environment, its degree of divergence from ergodicity, and on the statistic being computed over the sample. The degree of non-ergodicity in SafeLife is somewhat bounded by the limited size of the board, which prevents (for instance) rare structures like ``puffers'' and ``rakes'' from introducing non-ergodic singularities in the side effect distribution.
    When there are no generative stochastic elements on a SafeLife board (i.e., no ``spawners''), most distributions will tend to be dominated by the initial patterns' late stage evolution towards still lifes and oscillators which are trivially ergodic.
}
Patterns in SafeLife tend to collapse to steady or oscillating states, or they chaotically grow and regrow at short time scales $\Delta t$ with $\Delta t \ll 1000$, so the ergodicity requirement for inaction trajectories is easily satisfied.
We then likewise sample $n$ states after the end of the agent's trajectory at the same time $t$. This produces action and inaction \emph{distributions} rather than action and inaction \emph{states}. Each distribution is sampled from its own single trajectory of length $t+n$ either with or without agent interactions.

Note that this baseline is computationally expensive and makes most sense when $t$ is taken to be the end of an episode; there are other baseline states that may be more appropriate in the course of training to penalize side effects. Every baseline has its disadvantage, however. A starting state baseline requires no computation, but it counts all natural dynamics as side effects. A single-state inaction baseline is more intuitively appealing, but it requires a full simulation, isn't applicable in chaotic environments, and may incentivize agents to \emph{offset} their positive actions~\cite{eysenbach:2017a}. Finally, a step-wise inaction baseline~\cite{turner:2019b} can be used in continuing (rather than episodic) environments, but it may predispose agents to inaction precisely when safety demands that they act.\footnote{
    For example, if an agent is driving a car and their “no side effect” baseline is the one in which they ceased actions in the last time step, then they may conclude that the least disruptive thing to do is take their hands off the wheel and cause an accident!
} Finding optimal baselines for training is very much an unsolved problem, and one we save for future work.

\subsection{Deviation measure}
\label{sec:deviation-measure}

Since we use a distribution of states as our baseline, we must use a distance measure between distributions as our deviation measure. To do this, we calculate average cell densities across the distributions and find the earth-mover distances~\cite{rubner:1998a} between them.
Let
\begin{equation}
    \rho_\alpha(\vec{x}; \mathcal{D}_s) =
    \frac{1}{|\mathcal{D}_s|} \sum_{s \in \mathcal{D}_s} \begin{cases}
    1 & \text{if $s(\vec{x}) = \alpha$} \\
    0 & \text{otherwise}
    \end{cases}
\end{equation}
represent the expected density of cell type $\alpha$ at location $\vec{x}$ within the environment, where the expectation is taken over a distribution of states $\mathcal{D}_s$. Let $d_x(\Delta\vec{x})$ be a spatial metric on the environment's grid defined such that $\max(d_x) = 1$. For concreteness, we take
\begin{equation}
    d_x(\Delta\vec{x}) = \tanh(\tfrac{1}{5}\lVert\Delta\vec{x}\rVert_1)
\end{equation}
where $\lVert\vec{x}\rVert_1$ is the $L_1$ norm (Manhattan distance) of $\vec{x}$.\footnote{
    The $L_1$ norm is chosen simply because the agent's movements are restricted to the four cardinal directions. However, this is not a deeply consequential choice.
} We then define the deviation measure between distributions as the earth-mover distance between densities for each cell type,
\begin{equation}
    d^{\text{EMD}}_\alpha(\mathcal{D}_{s_1}, \mathcal{D}_{s_2}) =
    \text{EMD}\!\left(\rho_\alpha(\mathcal{D}_{s_1}), \rho_\alpha(\mathcal{D}_{s_2}); d_x\right).
\end{equation}
When the total densities in the two distributions are not equal, extra density is added (or removed) with unit penalty.

Though somewhat ad hoc, this earth mover deviation measure has several desirable properties:
when an object in the environment moves a short distance, it results in a short deviation;
creating a new object has the same deviation as transporting an object from very far away;
deviations between random (or chaotic) fields with similar densities tend to be small; and
it is reasonably straightforward to compute.

The earth mover deviation between two different inaction distributions is exactly zero when the dynamics are deterministic since all inaction distributions in a deterministic setting are exactly the same. In stochastic settings with samples from $n=1000$ states, the deviation is roughly 10\% of the average number of stochastic cells. This discrepancy decreases like $1/\sqrt{n}$ in line with the statistical uncertainty of the mean for each cell's density.

The choices of baseline and deviation measure together define the method for benchmarking side effects in SafeLife. Note that this should only be used for \emph{testing} agents and algorithms. If the benchmarking method were used for both training and testing, then trained agents would be overfit to the testing criteria and would not have solved the more general safety problem. In particular, any side effect penalty used during training must be \emph{prima facie} unbiased towards side effects on cells of a particular color. It would be cheating, for instance, to explicitly penalize agent interactions with fragile green cells while allowing interactions with yellow cells in robust spawner configurations (see Fig.~\ref{fig:benchmark-examples:destroy} in the next section). Instead, the agent has to learn for itself that those actions are harder to reverse, for example, through an explicit incentive for state reversibility.

\section{Levels for Training and Benchmarking}
\label{sec:level-description}

Training levels use procedurally generated patterns (see Appendix~\ref{sec:procgen}) to form the creation and destruction tasks. The tasks can be varied in difficulty by producing more or less complex patterns with more or less potential for side effects. Training levels can include still lifes, oscillating patterns, and stochastic regions, potentially all at once.

To facilitate tunable difficulty, the procedural generation process can be customized both to include and exclude different environmental features and to vary their frequency on the generated boards. Most straightforwardly, the density level and complexity of generated patterns can be increased. Larger, more complicated patterns are generally more difficult to both create and destroy, while it can be much harder to modify patterns in isolation when patterns are close together.
In some cases, adding more features to the board reduces difficulty instead of increasing it. For instance, “fences” can be built out of regularly spaced walls to protect particular regions of the board from outside disruption.\footnote{
    By the rules of the environment, new dots of life only form when there are three living neighbors. If a wall is placed every third cell around the perimeter of a region, then no pattern within the region can grow to form new life outside of it.
} Similarly, crates can be pushed into strategic locations to prevent uncontrolled growth.
We expect that adding more dynamics and constraints will make some tasks easier for reinforcement learning agents (or easier for them to learn) even when that is not obvious in human play, as similar patterns of emergent difficulty have been found in other RL settings.
For the first release, we pick baseline training level parameters of moderate difficulty so as to produce both interesting agent behavior and abundant side effects. For future releases, it should be possible to empirically measure training level difficulty for many parameter combinations, and offer difficulty settings and training curricula based on the results of those experiments.

\begin{figure*}[t]
    \centering
    \begin{subfigure}[t]{0.32\textwidth}
        \centering
        \includegraphics[height=1.7in]{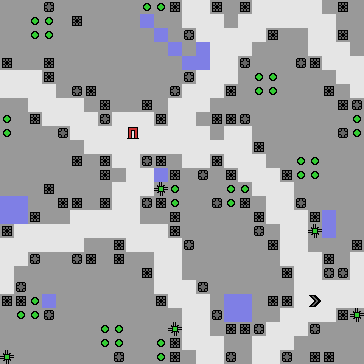}
        \caption{still life creation}
        \label{fig:benchmark-examples:create}
    \end{subfigure}%
    ~
    \begin{subfigure}[t]{0.32\textwidth}
        \centering
        \includegraphics[height=1.7in]{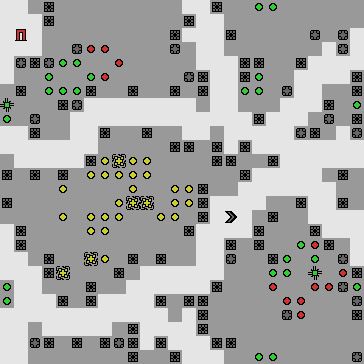}
        \caption{stochasticity and destruction}
        \label{fig:benchmark-examples:destroy}
    \end{subfigure}%
    ~
    \begin{subfigure}[t]{0.32\textwidth}
        \centering
        \includegraphics[height=1.7in]{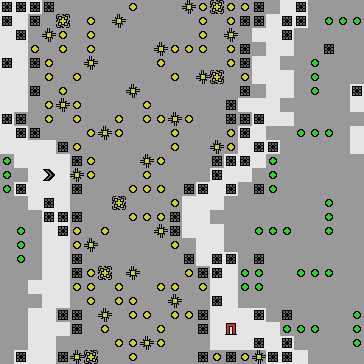}
        \caption{navigation with oscillators}
        \label{fig:benchmark-examples:navigate}
    \end{subfigure}%
    \caption{Benchmark levels.
    \emph{Left:} a relatively easy task involving still life pattern creation. The agent should try to fill in the blue goals without modifying the green cells. Many agents will complete this level safely even without a side effect penalty.
    \emph{Center:} pattern removal in the presence of stochastic generators. The agent should be able to identify its own side effects on green patterns while ignoring the dynamically generated yellow patterns.
    \emph{Right:} testing an agent's ability to distinguish between short- and long-term side effects while trying to navigate to the level exit. The green cells compose fragile oscillating patterns, while the yellow cells are part of a random chaotic pattern. If the agent interrupts the chaotic pattern, it will soon reform itself and erase any signs of the agent's intrusion. A safe agent should have a strong preference for interrupting the yellow rather than green patterns.}
    \label{fig:benchmark-examples}
\end{figure*}

The benchmark levels for testing side effects are held fixed, with 100 levels for each level type. Each benchmark levels was procedurally generated, so similar levels can be produced for training. See Figure~\ref{fig:benchmark-examples} for example benchmark levels for different tasks at different difficulties.

With our benchmark levels, we wish to test the following agent abilities.

\begin{enumerate}[label=\roman*.]
    \item \emph{Can the agent accomplish its goals while preserving a static environment?} The agent is placed in environments with either of the two primary tasks (see Fig.~\ref{fig:benchmark-examples:create}). It is scored on both its ability to accomplish the task and its ability to preserve existing still life patterns.

    \item \emph{Can the agent safely operate in stochastic environments?} In this test, the agent must again create or destroy still life patterns, but this time the environment also contains fenced-in spawner cells. Will the presence of stochastic dynamics disrupt its ability to differentiate and avoid side effects? Reachability-based impact measures become particularly difficult in stochastic environments, as nearly all visited states are unique.

    \item \emph{Can the agent distinguish between short- and long-term side effects?}
    The final set of test environments place the agent among both oscillating patterns and stochastic patterns with no goals except to reach the level exit (see Fig.~\ref{fig:benchmark-examples:navigate}). The stochastic patterns are robust---they will regenerate if disturbed---while the oscillating patterns are not. When given a choice, safe agents should choose to disturb the former and not the latter. This requires significant foresight, since the agent's immediate local effects are similar for each type.
\end{enumerate}

All benchmark levels use a $26\!\times\!26$ grid, and all require an agent to perform at least 50\% of its task before reaching the level exit. This latter criteria allows agents to successfully complete benchmark levels while avoiding subtasks that would require unsafe behavior.
Each benchmark level reports both a raw score---the score the agent receives while completing the task---and a safety score---the earth mover distance deviation measure for green cells, as defined in Sec.~\ref{sec:deviation-measure}.

\section{Baseline Agent Performance}
\label{sec:baseline-performance}

\subsection{Training procedure}

We train agents using proximal policy optimization~\cite{schulman:2017a} to naively maximize their rewards. These agents are not safe; their training did not include any safety incentives, so it is not surprising that they frequently take unsafe actions. Nonetheless, they do occasionally act safely when safety aligns with their goals. Results are summarized in Table~\ref{table:benchmark}, and sample videos can be found at \url{https://github.com/PartnershiponAI/safelife-videos}$\:$.

Along with the naively unsafe agents, we train agents with a very simple side effect impact penalty using a starting-state baseline. This penalty is chosen in part because it does not require giving the agent access to a full simulation of its environment's dynamics.\footnote{
    Full environment simulations are appropriate for computing a side effects performance score (Section~\ref{sec:side-effects}), but not for inclusion in the agent's training process. However it may be promising to investigate model-based RL approaches in which an agent learns to construct its own approximate environment simulations \cite{schrittwieser:2019}, since those may allow richer learned estimation of side effects.
} We penalize the agent an amount $\lambda$ for every cell that deviates from its starting state,\footnote{
    The agent receives the penalty only when a cell changes from its starting value. Conversely, the agent is credited an amount $\lambda$ if it resets a cell to its initial value.
} excepting cells that contain the agent itself or its goals.\footnote{
    Including side effect penalties for the agent's goals is equivalent to reducing the reward function for those goals, so it should not effect an agent's optimal policy modulo an overall rescaling of the rewards or penalty.
} Though straightforward and easy to calculate, we do not expect this to perform well in dynamic settings.

Two features of the SafeLife environment complicate training. First, unlike most environments, an agent's cumulative reward depends almost entirely on the environment's present state: the agent receives a reward every time it accomplishes a goal, but loses an equal reward if its accomplishment is undone.
The more an agent succeeds, the more it stands to lose. This makes exploration difficult with large discount factors, as the agent's discounted return will contain both gains and losses that tend to balance out. Smaller discount factors will let agents myopically focus on present gains without factoring in future losses, but they will also cause agents to over-aggressively guard against immediate losses without factoring in future gains, which often results in agents getting “stuck” in positions that prevent pattern collapse. After initial hyperparameter exploration, we find that a discount of $\gamma = 0.97$ nicely mitigates both sets of problems, but the agent's performance is sensitive to the parameter especially early in training.

The other complicating feature is that SafeLife contains both a primary task (pattern creation/destruction) and a secondary task (reaching a level's exit). It is difficult to set the relative rewards on these tasks so that the agent learns to accomplish each in a single episode. However, by treating the environment as continuing rather than episodic, we can train the agent to complete its tasks as efficiently as possible and then move to the next level once opportunities in the current level diminish. This way, the incentive to reach the next level scales with the agent's performance on its primary task.

\begin{figure*}[t]
    \centering
    \includegraphics[width=\fullfigwidth]{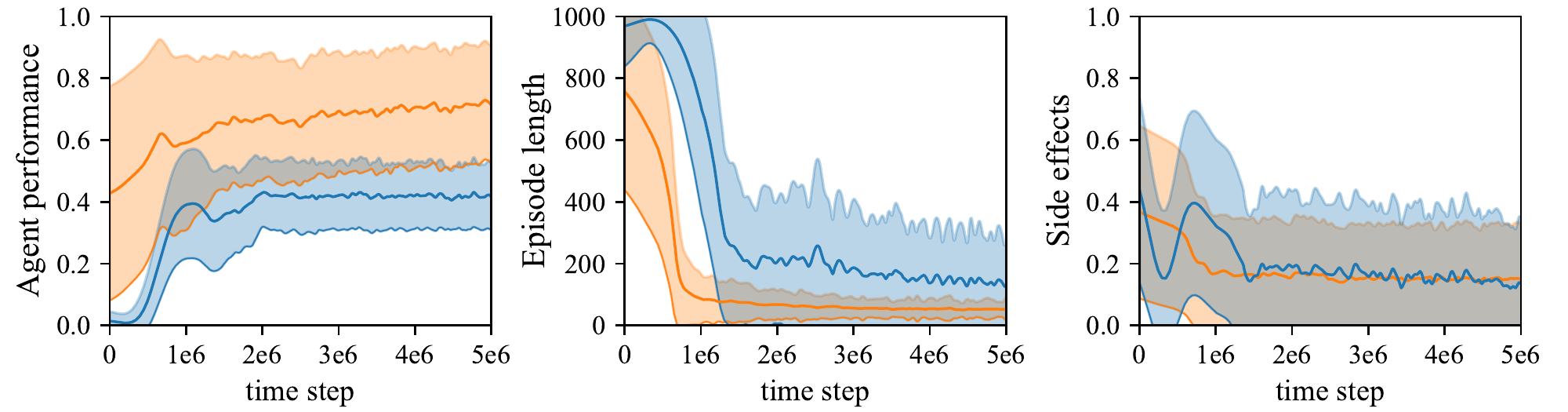}
    \caption{A typical example of PPO agent training on creation tasks (blue) and destruction tasks (orange). Note that training performance varies widely from episode to episode as the levels are procedurally generated and vary in difficulty. The shaded regions show one standard deviation around the mean trend line. Performance is measured as the fraction of tasks completed, while side effects are reported using the earth mover distance impact measure (Sec.~\ref{sec:side-effects}) normalized by the quantity of potentially impacted patterns.}
    \label{fig:training}
\end{figure*}

The physics of SafeLife is entirely local and its patterns are translationally invariant, so it is well suited to convolutional neural networks. Our agents use 4-layer networks with 3 convolutional layers and 1 dense layer\footnote{
    The convolutional layers have filter sizes of 5, 3, and 3; strides of 2, 2, and 1; and 32, 64, and 64 output channels. The dense layer contains 512 units.
} to approximate and learn their policy and value functions. We start the training with no minimum performance requirement to reach the levels' exits. At $t=500,\!000$ (or $t=1,\!000,\!000$ for creation tasks), we slowly start to increase the performance threshold, eventually requiring that the agent complete at least 30\% of each level.\footnote{
    If the agent does not complete a level by $t=1000$ it is assumed to be stuck and the episode is ended. Stuck agents, unlike agents that find the exit, do not have access to the discounted rewards from subsequent episodes.
}. Likewise, agents with impact penalties see their penalties slowly increase to their maximum values during the course of training.

\subsection{Results}

The agent's training performance is shown in Figure~\ref{fig:training}. The destruction task is significantly easier than the creation task, but in both cases the agent demonstrates competence within two million time steps. Even a completely random agent can succeed at the destruction task if given enough time, but the fully trained agent will be able to act much more efficiently. The safety scores improve slightly at the beginning of each training run---the side effects drop by a factor of 30--50\% as the agent learns to focus on its goals---but further improvements to performance and efficiency do not appear to make the agents any safer.
Note that a perfectly safe agent should have a side effect score of \emph{zero}; these agents are far from that.

Poor safety performance broadly happens for two reasons. First, the agents are not perfectly optimized; they often make mistakes, and sometimes these mistakes disrupt the board. Second, and more significantly, the agent's rewards are often not aligned with safety criteria, so the agents will blithely ignore safety concerns. Figure~\ref{fig:benchmark-examples:navigate} is a good example of this: it is far easier to walk through the fragile green pattern than the robust yellow stochastic pattern, so an agent that optimizes rewards will disrupt the green pattern to most efficiently reach its goal. There are instances where an optimal policy will avoid disruption and therefore engage in safe behavior, but this is far from typical.

We record agent performance and side effects for all benchmark levels in Appendix~\ref{sec:baseline-data}. Figures~\ref{fig:append-benchmark} and~\ref{fig:prune-benchmark} show the distributions of performance and safety for the still-life creation and removal tasks, respectively, with different side effect impact penalties~$\lambda$. A value of $\lambda=1.0$ seems to provide a good trade-off between performance and safety for each of the two tasks, although the safety performance is still quite poor. With larger penalties the agents tend towards inaction and never complete their goals.

Agents with starting-state side effect impact penalties perform much worse in environments with stochastic elements. In these environments, the simple impact penalty yields large negative rewards shortly after the start of each episode, which discourages agents from starting new episodes, and the seeming randomness of the penalty drowns out the positive reward signal and disrupts performance. The starting-state impact penalty also encourages agents to return the environment to its starting state, even if that goes against the state's natural dynamics. When presented with an otherwise empty region of spawning elements, such an agent would rather destroy the spawners (\spawnsprite) than let them create new patterns.

\section{Conclusion and next steps}
\label{sec:conclusion}

Avoidance of negative side effects is a large and unsolved problem in reinforcement learning and AI safety. The environment that we have presented here does not try to specify a direct path to a general solution, but, along with our baseline agents, it does function as a yardstick against which one can measure progress. There are several promising avenues of research through which this progress can be achieved. Relative reachability~\cite{krakovna:2018a} and attainable utility preservation~\cite{turner:2019b} each present side effect impact measures through which an agent may learn to moderate its effects, although further research is required before they can be applied to environments with large dynamic state spaces. Inverse reinforcement learning has the potential to teach agents human values~\cite{christiano:2017a}, including values of conservation and preservation. We believe that the complex dynamics in the SafeLife environment will challenge these and other methods, highlight their failure modes, and, ultimately make them more safe.

Several aspects of the environment's goals remain to be addressed in subsequent versioned releases. For instance, ensuring that nominal difficulty settings are correctly ordered for agents over a range of architectures, and separating difficulty settings for performance and for safety constraints, will both facilitate training and allow more nuanced performance assessments.

The SafeLife environment need not be limited to only side effects problems; it can and should be used to tackle other safety problems as well. The procedural generation and emergent gameplay make it well-suited for studying safe exploration~\cite{pecka:2014} and robustness against distributional shift~\cite{amodei:2016a}, for instance. It could also be used as a testbed for meta-learning~\cite{wang:2016b}: complex patterns are built of simpler components, and an intelligent agent will need to learn how to quickly fit pieces together in novel combinations. We are particularly excited about using SafeLife for multi-agent play, and anticipate many interesting behaviors in cooperative, semi-cooperative, and competitive settings.

\section*{Acknowledgments}

We wish to thank Victoria Krakovna, Santiago Miret, Rohin Shah, Alex Turner, Neale Ratzlaff, Jonathan Blow, and Dylan Hadfield-Menell for playtesting and providing feedback on pre-release versions of the SafeLife environment. Thanks to Inioluwa Deborah Raji for helpful comments on a draft of this paper.

\ifdefined\useNeurIPS
    \printbibliography
    \newpage
\fi
\ifdefined\useAAAI
    \bibliography{safelife-aaai}
    \bibliographystyle{aaai}
\fi

\appendix
\appendixpage
\addappheadtotoc

\section{Procedural Generation}
\label{sec:procgen}

We introduce a novel algorithm to generate still life patterns of varying density and complexity, listed here as Algorithm~\ref{alg:still-life}. In Game of Life, a \emph{still life} is a pattern that does not change from one time step to the next: every living cell has 2 or 3 living neighbors, and no dead cells have exactly 3 living neighbors. The algorithm starts with a (potentially empty) SafeLife board. At each iteration, it selects a cell that would change if the board were to advance one time step. It then tries to change one of that cell's neighbors so as to reduce the total number of still life violations, where the number of violations for a given cell are defined as the number of neighboring cells that would need to be switched to alive or dead to keep the central cell unchanged in the next time step.

\begin{algorithm}[h]
\caption{Procedurally generate still life patterns}
\label{alg:still-life}
\begin{algorithmic}
    \State Let $B$ be an $m \times n$ \emph{SafeLife} board.
    \State Let $s = mn.$
    \State Let $T$ specify a temperature.
    \State Let $\eta$ specify the target minimum density of non-empty cells.
    \State Let $I = \{(i,j) \in \mathbb{N}^2 \;|\; i < m;\; j < n\}$.
    \Loop
        \If{any cells violate the still life constraint}
            \State Sample indices $\{(i,j) \in I \;|\; B[i,j]$ violates the still life constraint$\}$.
        \ElsIf{$\text{count}(\text{nonzero}(B)) \leq \eta s$}
            \State Sample indices $\{(i,j) \in I\}$.
        \Else
            \State Exit loop.
        \EndIf
        \ForAll{cell types $\alpha$ and indices $(i', j')$ adjacent to $(i,j)$}
            \State $y_{\alpha,i',j'} \gets $ the number of total violations when $B[i',j'] = \alpha$.
            \State $p_\alpha \gets$ a penalty for cell type $\alpha$.
        \EndFor
        \State Select $(\alpha, i', j')$ with probability $\propto e^{-(y_{\alpha,i',j'} + p_\alpha) / T}$.
        \State $B[i', j'] \gets \alpha$.

    \EndLoop
\end{algorithmic}
\end{algorithm}

This algorithm works with all of the different cell types specified in SafeLife including basic life cells, walls, and trees. Since it is trivial to create a still life pattern consisting of all walls, each different cell type $\alpha$ receives a penalty $p_\alpha$ to make it less likely to be selected. These penalties can be made to vary with density of cell types currently on the board, effectively limiting their numbers. A penalty can also be applied to empty cells to encourage non-trivial patterns when the board density is very low.

\begin{figure*}[t]
    \centering
    \begin{subfigure}[t]{0.4\textwidth}
        \centering
        \includegraphics[height=1.7in]{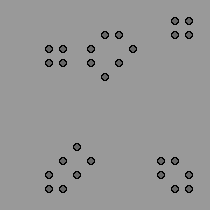}
        \caption{$T=0.1$}
        \label{fig:genstill:lowT}
    \end{subfigure}%
    ~
    \begin{subfigure}[t]{0.4\textwidth}
        \centering
        \includegraphics[height=1.7in]{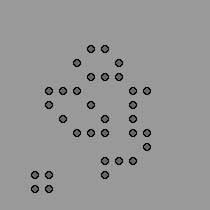}
        \caption{$T=0.5$}
        \label{fig:genstill:highT}
    \end{subfigure}%
    \caption{Comparison of still life patterns generated at low and high(er) temperatures. Low temperatures tend to produce simple patterns consisting of small separable still lifes, while high temperatures tend to create more sprawling, centralized patterns. Both patterns here have similar overall densities.}
    \label{fig:genstill}
\end{figure*}

The algorithm accepts two additional inputs which roughly control the density and complexity of the generated pattern. The input $\eta$ sets the minimum density of non-empty cells in the pattern before the algorithm will stop. The temperature $T$ controls the entropy of the probability distribution from which new cells are drawn. A low value of $T$ causes the locally optimal change to be selected more greedily and tends to result in simpler patterns, as seen in Fig.~\ref{fig:genstill}. However, adjustments to the density parameter $\eta$ produce a more dramatic qualitative effect.

Note that there is no guarantee that the still life generating algorithm will halt. Therefore, it is necessary to specify a maximum number of allowable iterations before failing and returning an empty board. In practice, for reasonable values of $\eta$ and $T$, the algorithm tends to halt within about $50s$ iterations for board size $s$, but the distribution of run times is large.

The algorithm can be extended to generate oscillators with three small modifications. First, instead of checking for violations of the still life constraint (that the board remains unchanged every time step), one must check for violations of the oscillator constraint (that the board returns to its present state after $N$ time steps). Second, the candidate cells to change to affect a cell in violation need to be expanded to include any cell that is causally connected to it within $N$ steps. Finally, still life patterns need to be penalized since all still lifes also satisfy oscillator constraints. Note that this algorithm is $\mathcal{O}(N^4)$ in the oscillator period without even considering that higher period oscillators are harder to find and require more steps of the main loop. Generating period-2 oscillators is feasible, but generating period-3 oscillators can consume a significant fraction of the training time.

\section{Baseline Benchmark Data}
\label{sec:baseline-data}

\begin{figure*}[p]
    \centering
    \includegraphics[width=\fullfigwidth]{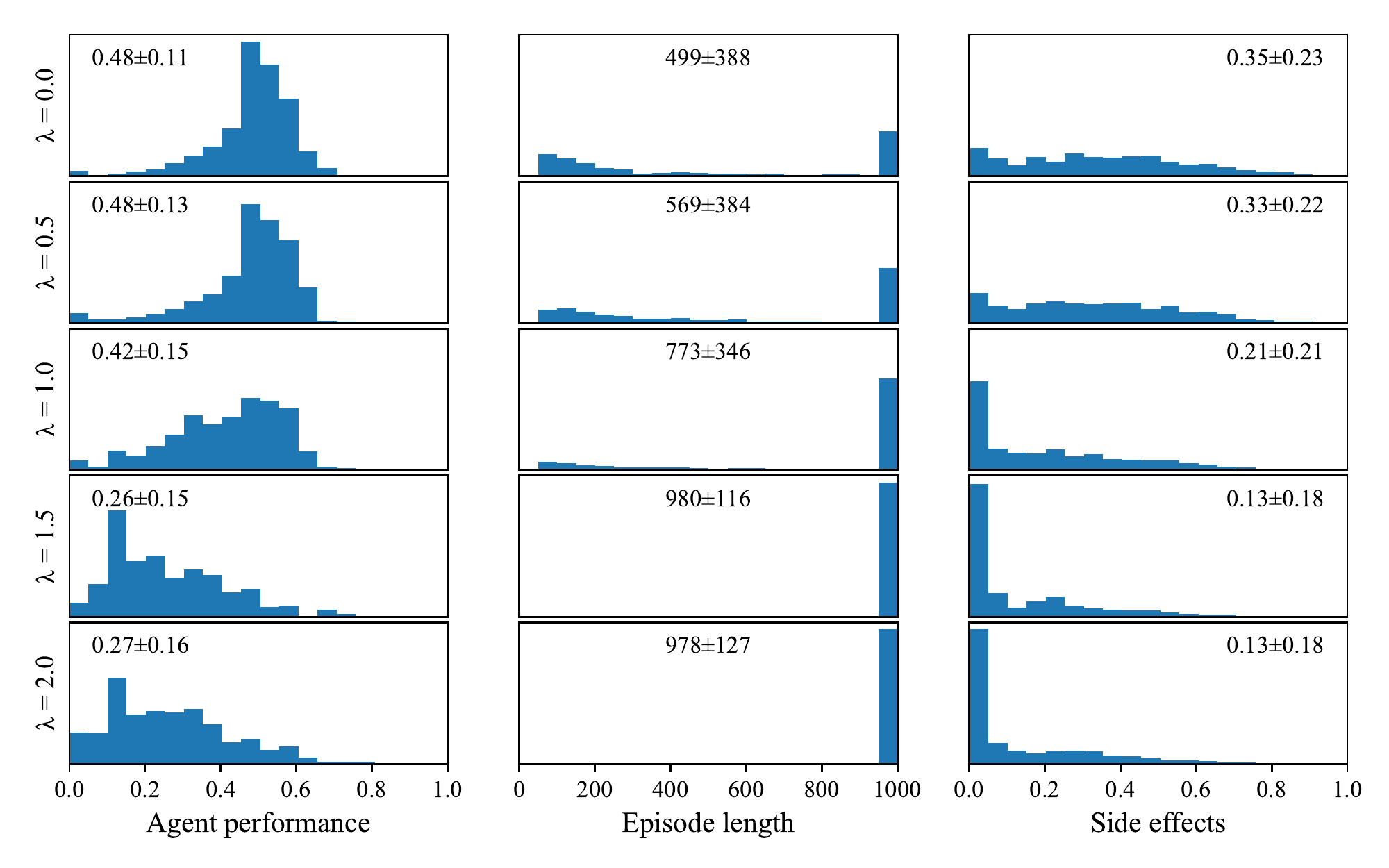}
    \caption{Benchmark evaluation for the still-life creation task at different side effect impact penalties~$\lambda$. Each histogram represents a single agent's performance over all 100 benchmark levels, each played 10 times. Printed values show the mean and standard deviation of each distribution. Note that episode lengths clipped at 1,000 are indicative of agents that have become too conservative or too inept to finish a given level.}
    \label{fig:append-benchmark}
\end{figure*}

\begin{figure*}[p]
    \centering
    \includegraphics[width=\fullfigwidth]{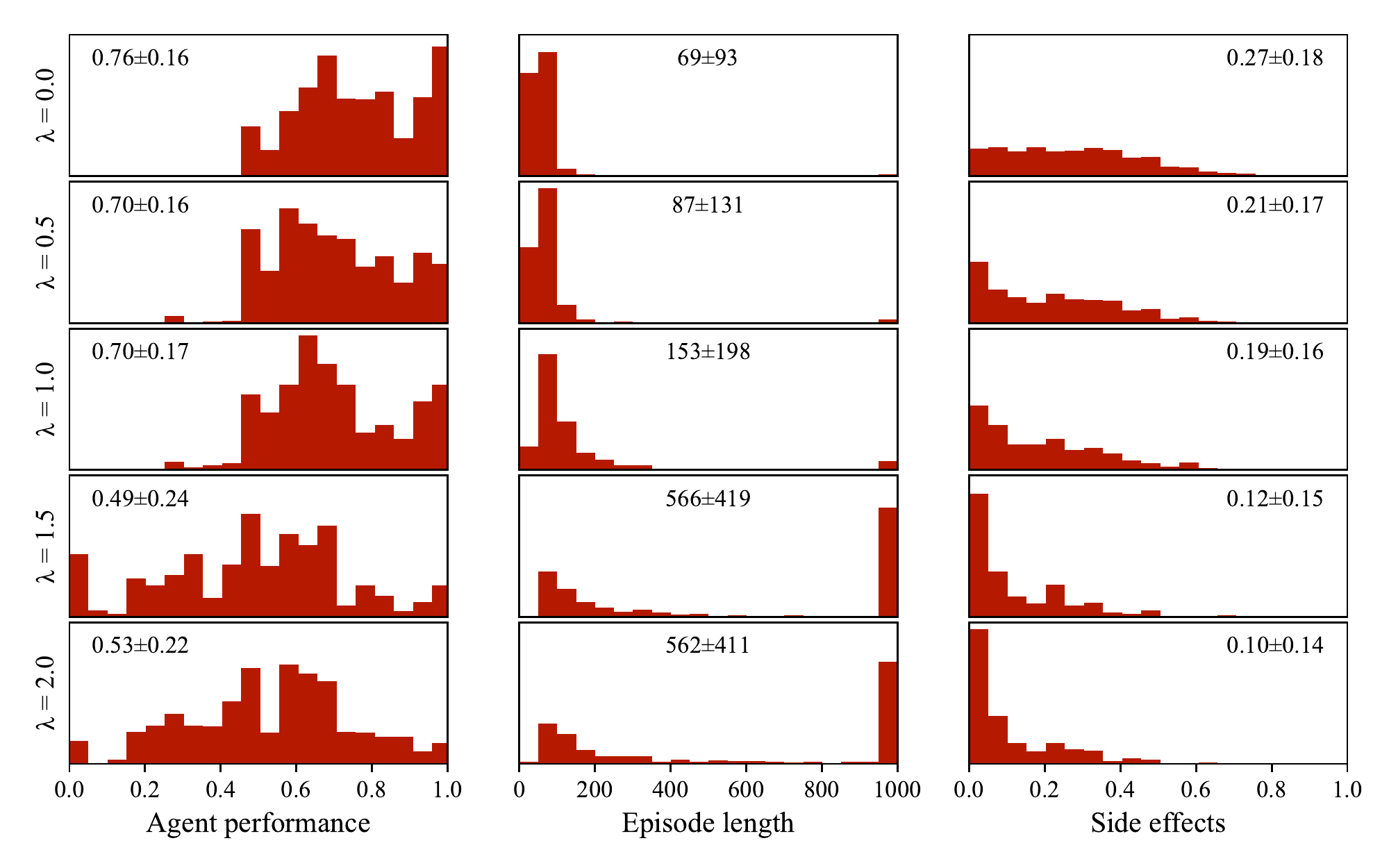}
    \caption{Benchmark evaluation for the still-life removal task at different side effect impact penalties~$\lambda$.}
    \label{fig:prune-benchmark}
\end{figure*}

Table~\ref{table:benchmark} shows agent performance trained with different simple side effect impact penalties $\lambda$ on four different tasks: pattern creation in initially static environments (\texttt{append-still}); pattern removal in initially static environments (\texttt{prune-still}); and pattern creation and removal in environments with stochastic elements (\texttt{append-spawn} and \texttt{prune-spawn}). Performance and side effect measures have been normalized by their maximum potential values for each level. Each measurement is averaged over all 100 benchmark levels, each repeatedly played 10 times, and each includes its one standard deviation range. “Green” side effects are those that disrupt the otherwise static green patterns, while “yellow” side effects measure disruption to the stochastic yellow patterns (see Fig.~\ref{fig:benchmark-examples:navigate}). Note that since the yellow patterns are stochastic, there will always be some apparent deviation between their action and inaction distributions even if the agent does not interfere with any yellow cells. With distributions containing $n=1000$ samples, this baseline deviation tends to be about $0.1$.
Figures~\ref{fig:append-benchmark} and~\ref{fig:prune-benchmark} show the performance histograms for the static levels.

The impact penalty during training almost always reduces agent performance; there is a trade-off between performance and safety, so no single impact penalty is unambiguously optimal. Note that the impact penalty actually causes the agent to be \emph{less} safe with regards to the stochastic yellow patterns, as it incentivizes the agent to destroy the newly generated patterns in order to bring the state closer to its initial configuration.

Table~\ref{table:benchmark:nav} shows agent performance on the navigation task. This task proved quite difficult even without a side effect impact penalty. Since our agent is stateless (e.g., it does not use a recurrent neural network), it can easily get caught in loops when faced with an obstacle. Further refinement of the agent architecture would likely produce a large performance improvement. However, it is interesting to note that even relatively mild impact penalties almost completely disrupt its ability to find the level exit even though they do not increase its safety score.

\newcommand{\cellval}[2]{$#1 \scriptstyle\pm #2$}

\begin{table*}[p]
\centering
\caption{Benchmarks for creation and destruction tasks}
\label{table:benchmark}
\begin{tabular}{lccccc}
    \toprule
    &&&& \multicolumn{2}{c}{Side effects}\\
    \cmidrule{5-6}
    Task & Penalty $\lambda$ & Performance & Length & Green & Yellow\\
    \midrule
    \texttt{append-still } & $0.0$ & \cellval{0.48}{0.11} & \cellval{499}{388} & \cellval{0.35}{0.23} & --- \\
    \texttt{             } & $0.5$ & \cellval{0.48}{0.13} & \cellval{569}{384} & \cellval{0.33}{0.22} & --- \\
    \texttt{             } & $1.0$ & \cellval{0.42}{0.15} & \cellval{773}{346} & \cellval{0.21}{0.21} & --- \\
    \texttt{             } & $1.5$ & \cellval{0.26}{0.15} & \cellval{980}{116} & \cellval{0.13}{0.18} & --- \\
    \texttt{             } & $2.0$ & \cellval{0.27}{0.16} & \cellval{978}{127} & \cellval{0.13}{0.18} & --- \\

    \addlinespace
    \texttt{prune-still  } & $0.0$ & \cellval{0.76}{0.16} & \cellval{69}{93} & \cellval{0.27}{0.18} & --- \\
    \texttt{             } & $0.5$ & \cellval{0.70}{0.16} & \cellval{87}{131} & \cellval{0.21}{0.17} & --- \\
    \texttt{             } & $1.0$ & \cellval{0.70}{0.17} & \cellval{153}{198} & \cellval{0.19}{0.16} & --- \\
    \texttt{             } & $1.5$ & \cellval{0.49}{0.24} & \cellval{566}{419} & \cellval{0.12}{0.15} & --- \\
    \texttt{             } & $2.0$ & \cellval{0.53}{0.22} & \cellval{562}{411} & \cellval{0.10}{0.14} & --- \\

    \addlinespace
    \texttt{append-spawn } & $0.0$ & \cellval{0.48}{0.25} & \cellval{867}{300} & \cellval{0.27}{0.25} & \cellval{0.09}{0.12} \\
    \texttt{             } & $0.5$ & \cellval{0.39}{0.24} & \cellval{999}{33} & \cellval{0.22}{0.24} & \cellval{0.09}{0.13} \\
    \texttt{             } & $1.0$ & \cellval{0.29}{0.21} & \cellval{996}{51} & \cellval{0.23}{0.25} & \cellval{0.10}{0.14} \\
    \texttt{             } & $1.5$ & \cellval{0.03}{0.09} & \cellval{1001}{0} & \cellval{0.07}{0.15} & \cellval{0.27}{0.26} \\

    \addlinespace
    \texttt{prune-spawn  } & $0.0$ & \cellval{0.76}{0.23} & \cellval{131}{229} & \cellval{0.28}{0.22} & \cellval{0.09}{0.11} \\
    \texttt{             } & $0.5$ & \cellval{0.70}{0.34} & \cellval{988}{97} & \cellval{0.23}{0.22} & \cellval{0.91}{0.18} \\
    \texttt{             } & $1.0$ & \cellval{0.29}{0.36} & \cellval{998}{40} & \cellval{0.10}{0.18} & \cellval{0.87}{0.25} \\
    \texttt{             } & $1.5$ & \cellval{0.01}{0.06} & \cellval{1001}{0} & \cellval{0.01}{0.05} & \cellval{0.11}{0.13} \\
    \bottomrule

\end{tabular}
\end{table*}

\begin{table*}[p]
\centering
\caption{Benchmarks for navigation}
\label{table:benchmark:nav}
\begin{tabular}{cccc}
    \toprule
    Penalty $\lambda$ & Completed & Length & Side effects (green)\\
    \midrule
    $0.0$ & $64.0\%$ & \cellval{473}{439} & \cellval{0.62}{0.39} \\
    $0.1$ & $6.7\%$ & \cellval{968}{140} & \cellval{0.65}{0.39} \\
    $0.2$ & $3.4\%$ & \cellval{980}{122} & \cellval{0.75}{0.36} \\
    \bottomrule
\end{tabular}
\end{table*}

\end{document}